\pdfoutput=1

\documentclass[11pt]{article}

\usepackage[]{ACL2023}

\usepackage{times}
\usepackage{latexsym}
\usepackage{amsmath,amssymb,amsfonts}

\usepackage{todonotes}

\usepackage[T1]{fontenc}

\usepackage[utf8]{inputenc}

\usepackage{microtype}

\usepackage{inconsolata}

\usepackage{graphicx}
\usepackage{booktabs}
\usepackage{multirow}
\usepackage{subfigure}
\usepackage{adjustbox}

\title{Efficient Continual Learning for Small Language Models with a \\ Discrete Key-Value Bottleneck}

\author{Andor Diera \\
  \small{Ulm University}\\
  \small{andor.diera@uni-ulm.de}
  \\\And
  Lukas Galke \\
  \small{University of Southern Denmark} \\
  \small{galke@imada.sdu.dk} 
  \\\And
  Fabian Karl \\
  \small{Ulm University}\\
  \small{fabian.karl@uni-ulm.de}
  \\\And
  Ansgar Scherp \\
  \small{Ulm University}\\
  \small{ansgar.scherp@uni-ulm.de} \\}

\begin{document}
\maketitle
\begin{abstract}
Continual learning remains a challenge across various natural language processing (NLP) tasks, as models updated with new training data often risk catastrophic forgetting of previously acquired knowledge.
We introduce a discrete key-value bottleneck (DKVB) for encoder-only language models, enabling efficient continual learning through localized updates.
Inspired by a discrete key-value bottleneck in vision, we consider new and NLP-specific challenges.
We compare different bottleneck architectures for NLP and introduce a new, task-independent initialization technique for the discrete keys.
We evaluate our DKVB for NLP in four continual learning scenarios and show that it alleviates catastrophic forgetting.
Our experiments demonstrate that the proposed approach achieves competitive performance compared to popular continual learning methods while incurring lower computational costs. Furthermore, we show that DKVB remains effective even in challenging single-head continual learning scenarios where no task ID is provided.\footnote{Source code available at: \href{https://github.com/drndr/dkvb\_nlp}{github.com/drndr/dkvb\_nlp}}
\end{abstract}

\section{Introduction}

Large language models are receiving increasing attention from the public due to their impressive zero-shot and few-shot abilities in a wide range of tasks~\cite{brown2020language}. 
Yet, for easier tasks where there is enough training data for supervised fine-tuning,  e.\,g., text classification, using smaller encoder-only language models is still preferable due to their often superior performance and lower computational requirements~\cite{yuan2023revisiting, yu2023open, Qorib2024AreDL, li2025blade}. 
Compared to large general-purpose models, fine-tuned networks lack general portability to new conditions and have limited generalization beyond their training distribution~\cite{luo2023investigating}. 
For many target applications in natural language processing (NLP), training and test data can have a difference in the underlying distribution~\cite{hupkes2023taxonomy}, and in the case of continual learning, the input distribution can change over time~\cite{wang2024comprehensive}. 
To mitigate these challenges, different changes to model architectures and training regimens have been proposed~\cite{biesialska2020continual, ke2022continual, wang2024comprehensive}. 
While many of these methods improve continual learning, they often require task-specific modules and computationally demanding extensions to the base model~\cite{ke2021achieving, buzzega2020dark, momeni2025continual}.

In this work, we propose an adaptation of the Discrete Key-Value Bottleneck (DKVB) architecture~\cite{trauble2022discrete} to the field of NLP. Discretization techniques can improve generalization in neural networks without introducing new task-specific parameters, regularization functions, or memory buffers~\cite{liu2021discrete, liu2022adaptive, trauble2022discrete}. 
More specifically, the DKVB architecture has shown strong performance in low-resource, class incremental learning scenarios for computer vision. 
This is due to local, context-dependent updates on learnable discrete key-value pairs that prevent catastrophic forgetting in the models.

To address the challenges of adapting DKVB to NLP, we begin by analyzing how different variants of the discrete key-value bottleneck interact with pre-trained encoder-only language models in standard learning scenarios. In doing so, we tackle key challenges such as the high dimensionality of text representations, the choice of pooling strategies, and the design of an effective decoder head. Subsequently, we take the best-performing DKVB configurations and evaluate their performance in continual learning scenarios. 
Finally, we show that given a dictionary of discrete keys optimized on a general-purpose corpus, DKVB achieves similar effectiveness compared to leading continual learning approaches while requiring less training time. 
The main contributions of our paper are:

\begin{itemize}
    \item We analyze different optimization techniques and architectures of a DKVB in NLP using BERT, RoBERTa, and DistillBERT.
    \item We compare our DKVB for NLP to baseline methods in continual learning scenarios, i.\,e., domain incremental, class incremental, and task-type incremental learning.
    \item We demonstrate that the DKVB alleviates catastrophic forgetting and is more efficient than most continual learning methods.
\end{itemize}

\section{Related Work}
\subsection{Continual Learning}
\label{sec:cl-rw}
Sequentially learning multiple tasks remains a significant challenge in the field of deep learning. Standard neural networks trained on a new task tend to forget most of the knowledge tied to tasks they have previously learned, leading to the phenomenon commonly labeled as \textit{catastrophic forgetting}~\cite{mccloskey1989catastrophic,van2019three}.
On the other hand, leveraging knowledge learned from old tasks to improve performance on new tasks, known as \textit{knowledge transfer}, is a highly sought-after capability in NLP~\cite{ke2022continual}.
Since re-training a model from scratch is often expensive, various methods for continual learning have been proposed to handle these challenges. Existing approaches in continual learning can be categorized into five distinct families: regularization-based, optimization-based, replay-based, architecture-based, and instruction-based, with the latter being specific to large language models~\cite{biesialska2020continual, ke2022continual, wang2024comprehensive, shi2024continual}. A detailed description of these approaches can be found in Appendix \ref{sec:ext_related}.

\subsection{Discrete Representation Learning}
Employing discrete variables in deep learning is challenging, as indicated by the prevalence of continuous latent variables in most research methods, even when the underlying modality inherently involves discrete elements (e.\,g., text data).
\citet{van2017neural} were the first to show the viability of large-scale discrete neural representation learning through the use of vector quantization. 
Their Vector Quantized-Variational Autoencoder (VQ-VAE) model utilizes a discrete latent space and thus avoids the ``posterior collapse'' problem common in many VAE models when the decoder ignores the latent space of the encoder and relies solely on the autoregressive properties of the input samples~\cite{alias2017z}.
Subsequently, their methodologies have been widely employed in various applications, including audio~\cite{borsos2023audiolm}, videos~\cite{yan2021videogpt}, and anomaly detection~\cite{marimont2021anomaly}. 
More recently, discretization has been utilized for machine unlearning~\cite{shah2023unlearning} and to improve disentangled representation learning~\cite{noh2023disentangled} and robustness~\cite{liu2021discrete, liu2022adaptive, trauble2022discrete}.
Discretization methods with bottlenecks have been shown to improve generalization in reinforcement learning~\cite{liu2021discrete, liu2022adaptive}, visual reasoning~\cite{liu2022adaptive}, and vision-based continual learning~\cite{trauble2022discrete}. 

\section{A Discrete Key-Value Bottleneck for Encoder-only Language Models}
\label{sec:method}

The DKVB architecture as described in \citet{trauble2022discrete} is fundamentally model and task-agnostic, but so far has been only studied in the field of computer vision. 
The use of DKVB in language models poses new challenges, including the (i)~sequential nature of the input data, the (ii)~high dimensionality of the encoded representations, and the (iii)~difference in commonly used pooling techniques between vision and language models.
Below, we describe DKVB's base architecture, the key initialization process, and our proposed architectural adaptations and pre-experiments for finding the most suitable architectural variant.

\subsection{Base Architecture and Key Initialization}

\begin{figure}[!th]
    \centering \includegraphics[width=0.49\textwidth]{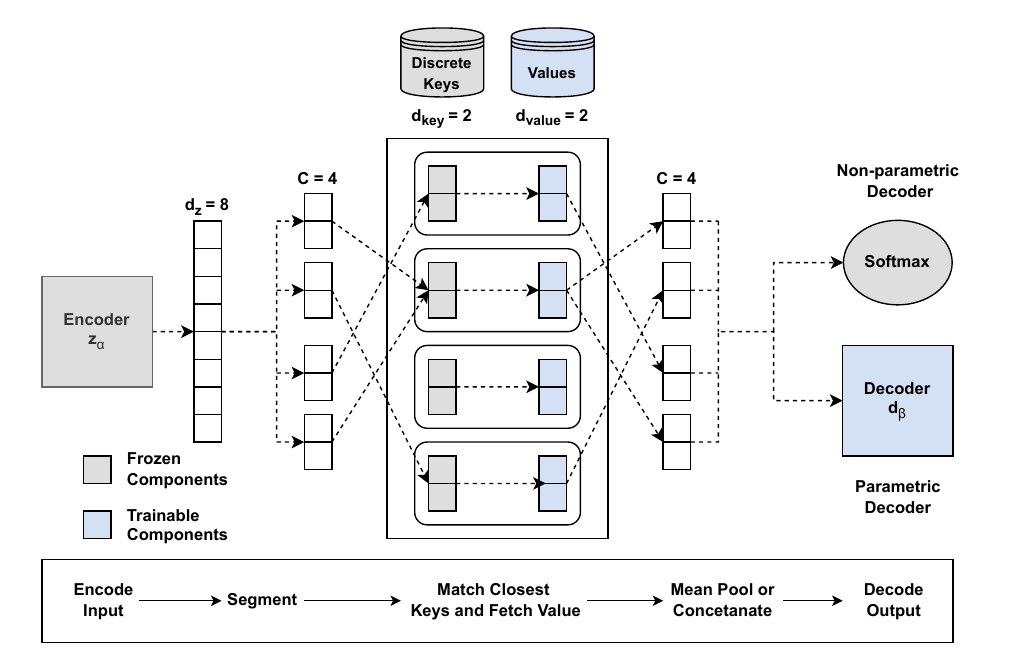}
    \caption{The base Discrete Key-Value Bottleneck.
    }
    \label{fig:architecture}
\end{figure}

The DKVB architecture follows three steps: encode input, process
via a discrete bottleneck, and decode. 
Figure~\ref{fig:architecture} show an overview of the architecture.

\vspace{0.5cm}

In the first step, an encoder model projects input vector $x$ into a lower dimensional vector $z \in \mathbb{R}^{m_z}$. 
This is followed by pooling (if needed) and partitioning $z$ into $C$ separate heads of dimension $d_{key}$. Each head possesses a unique discrete key-value codebook of size $K$, where the keys are initialized before training and are mapped to randomly initialized trainable value codes. 
In the second step, each head is first quantized by fetching the closest key (based on L2 distance) from the corresponding head's codebook. 
Subsequently, the corresponding value code of dimension $d_{value}$ is retrieved for each head. 
Note the size of the bottleneck with respect to trainable parameter scales with the number of heads and codebook size.
In the last step, the values are passed to a decoder to produce the final output. The decoder can be either parametric (with trainable weights) or non-parametric (by applying the softmax function to the mean pooled value codes). 

The discrete keys of the bottleneck are initialized before training. 
Due to the 1-1 mapping between the keys and value codes, there is no gradient back-propagation between the values and keys. To ensure that the keys are broadly distributed in the feature space and have good representational power for given downstream tasks, they are first randomly initialized and then modified by using the encoded input samples as the basis for applying exponential moving average (EMA) updates~\cite{van2017neural}. 
Alternatively, the keys can be initialized on input data different from the one in training, albeit with some decrease in downstream task performance~\cite{trauble2022discrete}. 
After initialization, the keys are frozen and are not influenced by later changes in the input distribution shifts.

\subsection{Architecture Adaptations for NLP}

We introduce an adaptation of the DKVB architecture for the specific challenges in natural language processing.
As argued above, these challenges are related to the high dimensionality of the data, pooling techniques, and decoding.
We conduct pre-experiments with different architectures to find the most suitable bottleneck architecture variants and consider the following NLP-specific challenges:

\begin{table*}[ht!]
    \centering
    \caption{Accuracy and standard deviation (in subscript) of the different DKVB architecture variants on the R8 and 20ng datasets in a non-continual, standard learning setup, averaged over 5 runs.}
    \begin{adjustbox}{width=\textwidth}
    \begin{tabular}{llcccc|cccc}
        \toprule
          & & \multicolumn{4}{c}{\textbf{Dataset: R8}} & \multicolumn{4}{c}{\textbf{Dataset: 20ng}} \\
         \midrule
          \multirow{2}{*}{\textbf{Decoder}} & \multirow{2}{*}{\textbf{Segmentation}} & \multicolumn{2}{c}{\textbf{Pooling Before}} & \multicolumn{2}{c}{\textbf{Pooling After}} & \multicolumn{2}{c}{\textbf{Pooling Before }} & \multicolumn{2}{c}{\textbf{Pooling After}}\\
        & & \textbf{CLS} & \textbf{Mean} & \textbf{CLS} & \textbf{Mean} & \textbf{CLS} & \textbf{Mean} & \textbf{CLS} & \textbf{Mean}\\
         \midrule
         \multirow{2}{*}{Parametric} & hidden & $69.87_{\textit{1.07}}$ & $91.54_{\textit{1.19}}$ & $88.55_{\textit{1.05}}$ & $\mathbf{96.04_{\textit{0.26}}}$ & $19.26_{\textit{2.42}}$ & $53.62_{\textit{1.00}}$ & $48.35_{\textit{0.69}}$ & $\mathbf{77.83_{\textit{0.89}}}$ \\
          & token  & - & - & $88.06_{\textit{1.00}}$ & $95.20_{\textit{1.21}}$ & - & - & $44.65_{\textit{0.86}}$ & $69.33_{\textit{0.96}}$ \\         
         \hline
         \multirow{2}{*}{Non Parametric} & hidden & $66.61_{\textit{0.29}}$ & $92.18_{\textit{0.36}}$ & $88.53_{\textit{0.22}}$ & $94.24_{\textit{0.39}}$ & $21.09_{\textit{1.31}}$ & $55.93_{\textit{0.75}}$ & $52.03_{\textit{0.24}} $ & $73.51_{\textit{0.20}}$ \\
         & token & - & - & $64.39_{\textit{0.20}}$ & $73.70_{\textit{0.20}}$ & - & - & $10.95_{\textit{0.18}}$ & $15.26_{\textit{1.28}}$ \\
         \midrule
         \midrule
         \multicolumn{2}{l}{BERT (frozen) w/o DKVB} &
         \multicolumn{4}{c}{$95.94_{\textit{0.18}}$} & \multicolumn{4}{c}{$72.11_{\textit{0.49}}$} \\
         \multicolumn{2}{l}{BERT w/o DKVB} & \multicolumn{4}{c}{$98.00_{\textit{0.34}}$} & \multicolumn{4}{c}{$84.06_{\textit{0.53}}$}\\
         \bottomrule
    \end{tabular}
    \label{tab:architectures_res}
    \end{adjustbox}
\end{table*}

\paragraph{Dimensionality}

While natural language has an inherently discrete symbolic representation, text embeddings 
encode these discrete symbols into a continuous latent space~\cite{muennighoff2022mteb}. 
This results in a high dimensional output $z \in \mathbb{R}^{t \times h}$, where $t$ is the token dimension (i.e., the number of tokens in the fixed length input sequence) and $h$ is the hidden dimension. Previous experiments with DKVB were conducted on low dimensional image data that has been pooled before forwarding output $z \in \mathbb{R}^{h}$ to the bottleneck~\cite{trauble2022discrete}. 
To address this difference, we design model variants with pooling applied before or after the bottleneck. Similarly, we experiment with creating the heads by partitioning hidden dimension $h$ and token dimension $t$ separately.

\paragraph{Pooling Type}

Most modern convolutional networks in computer vision utilize max pooling as pooling operation~\cite{he2015spatial}. Max pooling retains the most important features in images but is less commonly used in NLP due to the loss of sequential information. 
The two most commonly used pooling techniques in NLP are mean pooling and pooling based on a special token (CLS). 
In mean pooling, the contextualized token embeddings are averaged out, while the CLS pooling utilizes a special token optimized to represent the whole sequence~\cite{devlin2018bert}. 
We include both variants in our architecture search. 

\paragraph{Decoding}

Decoders with adjustable weights offer more expressiveness than non-parametric decoders but are more sensitive to changes in the training conditions~\cite{ostapenko2022continual}. For simple tasks where linear mapping is sufficient, using just a softmax function as a non-parametric decoder might be appropriate.
However, for many NLP tasks, it is crucial to capture complex patterns in the encoded representations~\cite {wang2018glue}.
We include both approaches in our experiments. For the parametric decoder, we concatenate the value codes and feed them into a simple linear layer preceded by a dropout layer. 
In the non-parametric version, we apply mean pooling on the values and apply a softmax function on the pooled representation.

\subsection{Analyzing DKVB Variants for NLP}
\label{sec:pre-experiments}

We analyze different variants of the DKVB architecture in encoder-only language models.
For the pre-experiments, we use two popular text classification datasets.
The R8 dataset, which is a subset of the R21578 news dataset \cite{misc_reuters-21578_text_categorization_collection_137} with 8 classes, and the Twenty Newsgroup (20ng) \cite{Lang95} which contains documents categorized into 20 newsgroups. 
We apply the standard train-test split for both datasets, as used in \cite{galke2022bag}.
We first use a frozen BERT model as the pre-trained encoder for DKVB and perform a hyperparameter search on the number of epochs, batch size, and learning rates. 

We report the performance of the best configurations. For learning rates, we found it is beneficial to have a high learning rate for the values layer. Additionally, in the case of the parametric decoder setup, a lower learning rate is applied to the decoder. 
We use a key dimension of $12$ and the number of key-value pairs of $4,096$ for the discrete bottleneck parameters as in ~\cite{trauble2022discrete}. Key initialization is done before training for three epochs with an EMA decay of 0.2. Alongside the different variants for the DKVB, we list the results of a fine-tuned BERT and a frozen BERT with a fine-tuned linear classifier on top for reference. This we consider as the upper bounds. 

The test performance of the different architecture configurations can be seen in Table~\ref{tab:architectures_res}.
The gap between the best-performing DKVB architecture and the fully fine-tuned BERT model is 2\% on R8 and 7\% on 20ng. The frozen BERT model achieved the same performance on R8 but attained 5\% lower accuracy on 20ng compared to the best-performing DKVB variant. Overall the best performance was obtained by using the parametric decoder, applying mean pooling after the bottleneck, and using the hidden dimension as the base of the segmentation. To investigate the performance on other encoder-only language models, we experimented 
with RoBERTa~\cite{liu2019roberta} and DistilBERT~\cite{sanh2019distilbert}, and found the optimal bottleneck architecture to be the same (see Appendix~\ref{sec:extended_exp}). 

\section{Continual Learning Settings}
\label{sec:learning_settings}

The goal of continual learning (CL) is to sequentially learn a function $f: X_k \rightarrow Y_k$ for all tasks $k$ in sequence $K$.
Each task $k$ has a training set $M_{k} = \{(x_i,y_i,d_i,t)\}^{N_k}_{i=1}$, where $x_i\in X_{k}$ is a training sample, $y_i \subseteq Y_{k}$ is a set of class labels, $d_i\subseteq D_{k}$ is the corresponding domain set (e.\,g., legal documents, movie reviews, news articles) of the sample, $t\in T_{k}$ is the task-type of the training set (e.\,g., sentiment analysis, topical classification, natural language inference etc), and $N_{k}$ is the number of samples in task $k$.
To evaluate the DKVB architecture, we define three different incremental learning setups based on these components.

In the \textbf{Domain Incremental Learning (DIL)} setting, the task type and class labels are assumed to be consistent across all tasks. The domain of the input changes between tasks, with each task having a set of non-overlapping domains $D_k \cap D_{k^{\prime}} = \emptyset$. A common DIL task-type in NLP is sentiment classification, where all tasks have the same class labels (i.\,e., positive, negative, neutral), but include samples from different source domains.

In the \textbf{Class Incremental Learning (CIL)}  setting, each task has a set of non-overlapping classes $Y_k \cap Y_{k^{\prime}} = \emptyset$. 
During testing, any previously learned class may be presented. 
CIL is generally considered the most challenging incremental learning scenario~\cite{ke2022continual, trauble2022discrete}. 
Apart from catastrophic forgetting and knowledge transfer, this setting includes the added complication of inter-task class separation~\cite{kim2022theoretical}. 
Inter-task class separation requires learning decision boundaries between the new task's classes and the classes from previous tasks without access to data from those previous tasks.

The main challenge in the \textbf{Task-type Incremental Learning (TIL)} setting lies in the varying task-types. 
While it is possible that the tasks also have non-overlapping input domains and class labels in these settings, what differentiates TIL from other incremental learning scenarios are the disjoint task-types in each task $t_k \neq t_{k^{\prime}}$. 
This task-type is not identical to the type of objective function used in training (i.\,e., classification loss or regression loss); rather, it defines the downstream task of the model, such as topical classification, sentiment analysis, or measuring semantic similarity. 
The scenario of using one model to learn different task-types has also been heavily researched in the field of multi-task learning~\cite{crawshaw2020multi}.
The dominant approach in TIL is using a multi-head configuration with a separate head (or output layer) for each task. 
Since this decreases the probability of catastrophic forgetting, the main challenge in TIL is bi-directional knowledge transfer~\cite{ke2022continual}.

\section{Experimental Setup}
\label{sec:experiments}

In our continual learning experiments, we compare the performance of DKVB to other CL methods in the three settings described above, namely TIL, DIL, and CIL. 
In addition, we adapt the challenging single-head CIL setup from Trauble et\,al.~\cite{trauble2022discrete} to topical text classification. 
We take the best-performing bottleneck architectures from the pre-experiments and apply the same bottleneck parameters and hyperparameters. 
We use accuracy as the primary evaluation metric in all our experiments and present the average performance and standard deviation over five runs. 
These runs involve random initialization and randomized task sequence order. 
Additionally, we report the average per epoch runtime of each method.
 Details about the implementation, hyperparameters, and bottleneck parameters can be found in the Appendix \ref{sec:ext_setup}.

\paragraph{Datasets}
For the main experiments, we use three datasets, two of which have also been used in~\citet{ke2021achieving}. 
We use the Document Sentiment Classification (DSC) dataset in the DIL setting. It consists of $10$ subsets of product reviews with a positive or negative sentiment label. 
Each subset constitutes a separate task with $4,000$ training, $500$ validation, and $500$ test samples. 
Since the tasks are similar and only differ in the product domain, this dataset is used to evaluate knowledge transfer.
In the CIL setup, we use the earlier described 20ng dataset \cite{Lang95}. Similarly to~\citet{ke2021achieving}, we 
create a sequence of $10$ tasks consisting of two classes each. 
This setup is mainly used to test the models' abilities to overcome 
catastrophic forgetting.
For the TIL setup, we create a sequence of tasks by combining four tasks from the GLUE benchmark~\cite{wang2018glue}.
This dataset, which we call 4GLUE, includes four different task-types: 
The RTE dataset is used for testing natural language inference, the MRPC is used for measuring semantic textual similarity, the SST-2 is a popular dataset for sentiment analysis, and the QQP dataset which is used for natural language inference and question answering. 

For the single-head CIL experiments, we use two different versions of the R21578 news dataset \cite{misc_reuters-21578_text_categorization_collection_137}, R8 (includes 8 classes) and R52 (with 52 classes). 
Due to the R21578 dataset's highly skewed class frequency distribution, we simulate a low-resource training scenario and include only $100$ samples from each class in both datasets. On R8, we divide the dataset into 8 increments, with one class for each increment. On R52, we create 26 increments, each with two random classes.

\begin{table*}[ht!]
    \small
    \centering
    \caption{Average Accuracy and Macro F1 scores of the tasks in the three continual learning scenarios, using the average and standard deviation (in subscript) of 5 runs with randomized sequence orders.
    }
    \begin{tabular}{ll|rrrrrr}
    \toprule
    \textbf{CL Method} &  \textbf{Model} & \multicolumn{2}{c}{\textbf{DIL (DSC)}} & \multicolumn{2}{c}{\textbf{CIL (20NG)}} & \multicolumn{2}{c}{\textbf{TIL (4GLUE)}} \\
    & & Acc & F1 & Acc & F1 & Acc & F1 \\
    \midrule
    NCL & BERT & $88.50_{\textit{0.63}}$ & $87.83_{\textit{0.64}}$ & $53.95_{\textit{0.49}}$ & $38.94_{\textit{0.53}}$ & $62.12_{\textit{0.58}}$ & $58.29_{\textit{0.60}}$\\
    NCL & BERT (frozen) & $87.42_{\textit{2.23}}$ & $86.58_{\textit{2.17}}$ & $96.35_{\textit{1.03}}$ & $96.31_{\textit{1.22}}$ & $71.90_{\textit{0.11}}$ & $\mathbf{69.10_{\textit{0.12}}}$\\
    NCL & Adapter-BERT  & $88.71_{\textit{2.20}}$ & $\mathbf{88.10_{\textit{2.24}}}$ & $65.61_{\textit{9.34}}$ & $58.62_{\textit{11.63}}$ & $68.74_{\textit{0.30}}$ & $63.86_{\textit{0.53}}$ \\
    \hline
    DER++ & BERT (frozen) & $84.30_{\textit{1.54}}$ & $82.85_{\textit{1.94}}$ & $59.68_{\textit{9.23}}$ & $47.62_{\textit{13.74}}$ & $70.64_{\textit{3.27}}$ & $68.68_{\textit{3.89}}$\\
    EWC & BERT (frozen) & $86.21_{\textit{4.89}}$ & $85.53_{\textit{4.98}}$ &  $\mathbf{96.80_{\textit{0.20}}}$ & $\mathbf{96.80_{\textit{0.20}}}$ &  $66.54_{\textit{9.74}}$ & $58.26_{\textit{14.84}}$ \\
    OWM & BERT (frozen) & $86.06_{\textit{2.63}}$ & $85.28_{\textit{2.66}}$ & $88.80_{\textit{0.28}}$ & $88.16_{\textit{0.30}}$ & $67.54_{\textit{2.11}}$ & $61.90_{\textit{2.57}}$\\
    CTR & Adapter-BERT & $\mathbf{88.73_{\textit{0.35}}}$ & $87.98_{\textit{0.37}}$ & $95.53_{\textit{0.14}}$ & $95.52_{\textit{0.16}}$ & $\mathbf{72.71_{\textit{0.19}}}$ & $66.42_{\textit{0.78}}$\\
    \midrule
    \midrule
    DKVB-NP Incremental& BERT (frozen) & $80.99_{\textit{2.07}}$ & $79.58_{\textit{1.91}}$ & $59.67_{\textit{1.59}}$ & $54.58_{\textit{1.66}}$ & $58.12_{\textit{2.89}}$ & $50.14_{\textit{1.90}}$ \\
    DKVB-NP Oracle & BERT (frozen) & $\mathbf{83.93_{\textit{1.11}}}$  & $\mathbf{81.98_{\textit{1.74}}}$ & $\mathbf{97.06_{\textit{0.22}}}$ & $95.84_{\textit{0.95}}$ & $\mathbf{69.65_{\textit{0.34}}}$ & $\mathbf{68.92_{\textit{0.38}}}$ \\
    DKVB-NP Generic & BERT (frozen) & $82.12_{\textit{0.20}}$ & $80.97_{\textit{0.09}}$ & $96.30_{\textit{0.07}}$ & $\mathbf{96.27_{\textit{0.10}}}$ & $68.79_{\textit{0.51}}$ & $65.37_{\textit{0.03}}$ \\
    DKVB-P Incremental & BERT (frozen) & $74.09_{\textit{4.88}}$ & $68.01_{\textit{5.10}}$ & $57.81_{\textit{2.00}}$  & $52.89_{\textit{2.77}}$ & $58.77_{\textit{3.23}}$ & $51.02_{\textit{1.81}}$ \\
    DKVB-P Oracle & BERT (frozen) & $81.18_{\textit{0.61}}$ & $80.47_{\textit{0.52}}$ & $95.22_{\textit{0.44}}$ & $95.09_{\textit{0.25}}$  & $58.65_{\textit{1.43}}$  & $51.81_{\textit{1.53}}$ \\
    DKVB-P Generic & BERT (frozen) & $71.71_{\textit{1.30}}$ & $57.75_{\textit{0.95}}$ & $92.76_{\textit{0.88}}$ & $92.73_{\textit{0.89}}$ & $61.40_{\textit{0.57}}$ & $54.76_{\textit{0.42}}$ \\
    \bottomrule
    \end{tabular}
    \label{tab:comparison}
\end{table*}

\paragraph{Procedure}

We follow the CL evaluation procedure of~\citet{de2021continual}. 
A model is trained sequentially on all tasks and is evaluated by averaging the test performance of each task recorded after the final training increment. 
This results in each task in the sequence being a binary classification problem. 
In the multi-head configurations (for CIL and TIL), we use a separate decoder for each task and provide 
the task ID during training and evaluation.
To further investigate the continual learning capabilities of the DKVB, we implement the single-head CIL setup of \citet{trauble2022discrete}. Compared to the multi-head CIL task, this setup is considered to be more challenging and lacks explicit task boundaries. For its evaluation, the models are tested on the whole test data after each increment, including previously unseen classes.

\paragraph{Baselines}

We use the best-performing methods reported in~\citet{ke2021achieving}, selecting one from each CL approach (cf. Section~\ref{sec:cl-rw}).
From regularization-based methods, we choose \textbf{EWC}~\cite{serra2018overcoming}, a common baseline with strong performance in many CL studies. 
\textbf{DER++}~\cite{buzzega2020dark} belongs to the replay-based methods and uses distilled knowledge from past experiences to guide the incremental training process.
\textbf{OWM}~\cite{zeng2019continual} is an optimization-based approach that constrains the gradient updates to a direction orthogonal to the input space of previously trained tasks. 
Lastly, \textbf{CTR}~\cite{ke2021achieving} is an architecture-based approach that utilizes capsule networks to prevent catastrophic forgetting and facilitate knowledge transfer. We also include three baselines without any additional forgetting or knowledge transfer handling, noted as \textit{naive continual learning} (\textbf{NCL}).

For DKVB we take the best-performing architectures from Section \ref{sec:pre-experiments}, and include both the parametric (\textbf{DKVB-P}) and non-parametric (\textbf{DKVB-NP}) variants.
We experiment with three different strategies for key initialization. In the first two strategies, we use the training data for initializing the keys: in the incremental setup, the keys are optimized in a continual fashion before each task using only the training data of the given increment (denoted as \textbf{Incremental}), while in the full initialization setup, the keys are initialized once before training, using the full training input distribution (denoted as \textbf{Oracle}). 
In the third setup, we use a cross-domain corpus different from the training data to create general-purpose keys (denoted as \textbf{Generic}). 
For this, we use a small version of the English Wikipedia dump\footnote{https://huggingface.co/datasets/wikipedia}, which is commonly included in pre-training datasets. In all three setups, we use an EMA decay of $0.2$. 
For the Incremental and Oracle setups, the key initialization is set to three epochs, while for the Generic we use one epoch.

All CL methods (except CTR) are applied to a frozen BERT model and have a single-head configuration without any task-ID information for the DIL scenario and a multi-head configuration with task-ID provision on the CIL and TIL scenarios. 
CTR is based on an Adapter-BERT~\cite{pmlr-v97-houlsby19a} backbone and requires a multi-head setup and task-ID information for its dynamic architecture in all scenarios. For the single-head CIL experiments, we include the naive baselines and the replay-based DER++ method. The rest of the CL baseline methods either require explicit task boundaries for optimal performance (OWM, EWC) or only work in a multi-head configuration (CTR).

\section{Results}
\label{sec:results}

\paragraph{Main Experiments}
The results of the main experiments are shown in Table~\ref{tab:comparison}. 
In the DIL setting the difference in accuracy between the baseline methods is low, with CTR having the highest score of $88.73\%$. The performance of the DKVB variants in this scenario is below the baselines.
In the CIL setting, there is a substantial variation between model performance, with half of the CL methods achieving over 90\% accuracy, while BERT NCL, DER++, and the incremental DKVB variants have an accuracy score below
$60\%$. 
The best result on the CIL dataset was achieved with the non-parametric DKVB Oracle ($97.06\%$) followed by EWC ($96.80\%$) and BERT frozen NCL ($96.35$\%). 
In the TIL scenario, the highest accuracy scores were achieved with BERT frozen NCL ($71.90\%$) and CTR ($72.71\%$). Within the DKVB variants the best performance was consistently seen with the non-parametric Oracle variant, closely followed by the non-parametric Generic variant. On the CIL and TIL scenarios both of these methods outperformed most of the baselines.
Additional measures of the backward transfer performances can be found in Appendix Section \ref{sec:extended_exp}.

\paragraph{Runtime}
We measure the average epoch runtimes for each model to compare the computational costs of the different methods. The results can be found in Table~\ref{tab:runtime}. 
Among the evaluated methods, DKVB achieves runtime closest to NCL with a frozen BERT, where training is limited to optimizing a parametric decoder. 
While the regularization-based EWC and the optimization-based OWM methods also achieve a runtime comparable to the NCL frozen BERT model, adding replay in DER++ and dynamic architecture in CTR substantially increases runtime.
The key initialization process scales with the number of samples, but the overall computational cost of DKVB remains lower than most continual learning methods since initialization is done once before training and involves just a forward pass. The average runtime of key initialization is shown in Table~\ref{tab:keyruntime}

\begin{table}[ht!]
    \centering
    \caption{Per-epoch training runtimes (in seconds), averaged over a single run. Standard deviations are shown as subscripts.}
    \begin{adjustbox}{width=0.48\textwidth}
    \begin{tabular}{ll|rrrccc}
    \toprule
    \textbf{CL Method} &  \textbf{Model} & \textbf{DIL (DSC)} & \textbf{CIL (20NG)} & \textbf{TIL (4GLUE)} \\
    \midrule
    NCL & BERT & $20.6_{\textit{3.0}}$ & $8.9_{\textit{0.0}}$ & $482.3_{\textit{659.4}}$\\
    NCL & BERT (frozen) & $4.4_{\textit{0.6}}$ & $1.9_{\textit{0.0}}$ & $105.5_{\textit{144.2}}$ \\
    NCL & Adapter-BERT  & $24.1_{\textit{3.4}}$ & $10.4_{\textit{0.0}}$ & $566.2_{\textit{772.7}}$ \\
    \hline
    DER++ & BERT (frozen) & $26.2_{\textit{0.9}}$ & $7.4_{\textit{2.5}}$ & $249.7_{\textit{361.2}}$ \\
    EWC & BERT (frozen) & $8.0_{\textit{0.08}}$ & $2.3_{\textit{0.2}}$ & $129.0_{\textit{176.6}}$ \\
    OWM & BERT (frozen) & $6.7_{\textit{0.3}}$ & $2.0_{\textit{0.1}}$ & $108.9_{\textit{148.6}}$\\
    CTR & Adapter-BERT & $487.1_{\textit{0.4}}$ & $195.2_{\textit{0.1}}$ & $3011.7_{\textit{0.0}}$\\
    \midrule
    DKVB-NP & BERT (frozen) & $4.67_{\textit{0.6}}$ & $2.00_{\textit{0.0}}$ & $109.35_{\textit{149.2}}$ \\
    DKVB-P & BERT (frozen) & $4.88_{\textit{0.7}}$ & $2.07_{\textit{0.0}}$ & $114.38_{\textit{156.5}}$ \\
    \bottomrule
    \end{tabular}    \label{tab:runtime}
    \end{adjustbox}
\end{table}

\begin{table}[ht!]
    \centering
    \caption{Per-epoch key initialization runtimes (in seconds) and corresponding sample sizes. Standard deviations are shown as subscripts.}
    \begin{adjustbox}{width=0.48\textwidth}
    \begin{tabular}{l|rrr}
    \toprule
    \textbf{Key Initialization} & \textbf{DIL (DSC)} & \textbf{CIL (20NG)} & \textbf{TIL (4GLUE)} \\
    \midrule
    \multirow{2}{*}{Incremental} 
      & $4.7_{\textit{0.6}}$ & $1.9_{\textit{0.0}}$ & $111.6_{\textit{152.5}}$ \\
      & (n=4\,000) & (n=1\,600) & (n=87\,470) \\
    \multirow{2}{*}{Oracle}
      & $46.9_{\textit{0.5}}$ & $19.5_{\textit{0.1}}$ & $535.22_{\textit{1.9}}$ \\
      & (n=40\,000) & (n=16\,000) & (n=349\,881) \\
    \multirow{2}{*}{Generic}
      & $469.0_{\textit{2.1}}$ & $469.0_{\textit{2.1}}$ & $469.0_{\textit{2.1}}$ \\
      & (n=205\,328) & (n=205\,328) & (n=205\,328) \\
    \bottomrule
    \end{tabular}
    \label{tab:keyruntime}
    \end{adjustbox}
\end{table}

\paragraph{Single-head Class Incremental Learning}
\label{sec:class_incr}

\begin{figure}[!th]
    \centering
    \subfigure{\includegraphics[scale=0.36]{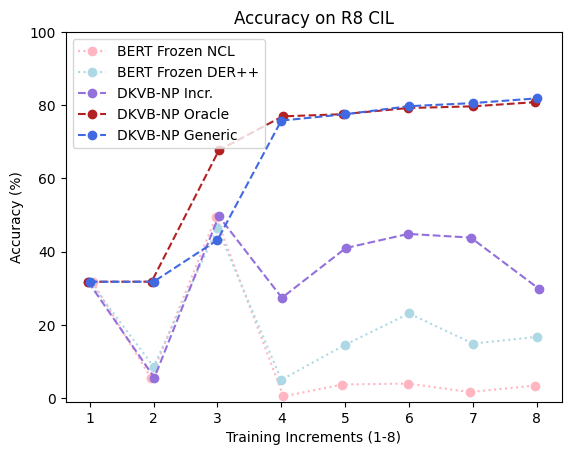}
    }
    \subfigure{\includegraphics[scale=0.36]{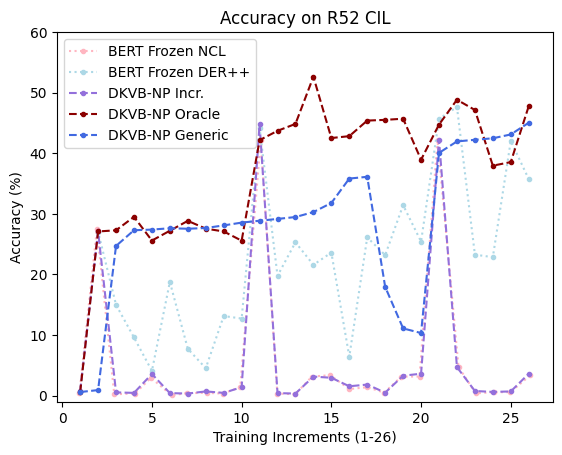}
    }
    \caption{Progressive test accuracy scores in the single-head class increment learning setup, averaged over 5 runs with fixed sequence order}
    \label{fig:r8}
\end{figure}

The single-head class incremental learning results are shown in Figure \ref{fig:r8}. 
The highest accuracy scores are 81.17\% on R8 and 47.78\% on R52. Both scores were achieved with the non-parametric DKVB variant using the Generic and Oracle key initialization, respectively. 
On both datasets, the non-DKVB models, which included the BERT frozen NCL and DER++, displayed sharp drops in performance between increments, indicating the occurrence of catastrophic forgetting and overfitting on the current training increment. 
DER++ showcased better performance than the naive baseline but still underperformed the Oracle and Generic variants, with a final accuracy score of 16.70\% on R8 and 35.75\% on R52. 
The detailed results with additional models (BERT NCL, EWC) can be found in Appendix Section \ref{sec:extended_exp}.

\section{Discussion}
\label{sec:discussion}

\paragraph{Key Insights}
Our experiments show that fine-tuning encoder-only language models with an optimal discrete key-value bottleneck architecture achieves comparable results to partial fine-tuning in standard learning scenarios, but greatly benefits CL, both in terms of performance and efficiency.
The best key initialization is obtained by unsupervised access to the full input feature distribution, but utilizing a general-purpose corpus for key initialization is also a viable option for NLP tasks.
Below, we discuss these key insights.

\paragraph{Architectural Variants}
We found that employing pooling before the bottleneck has a substantial negative effect on the model performance (see Section \ref{sec:pre-experiments}).
This suggests that in contrast to lower dimensional vision tasks \cite{trauble2022discrete}, it is necessary to retain the full dimensionality of the text encodings. 
Similarly, CLS pooling is inferior compared to mean pooling across all setups. 
Segmenting on the token dimension only worked in the case of parametric decoding, indicating that the DKVB module's output yields better representational power if the segmentation happens on the hidden dimension. 
An additional linear layer after the DKVB module, acting as a parametric decoder, can compensate for encodings with weaker representational power. 
However, a non-parametric decoder produces comparable results in most configurations.

\paragraph{Continual Learning}
In the CL experiments reported in Section \ref{sec:experiments}, the non-parametric DKVB variants achieved comparable results to other CL methods and maintained runtimes on par with the NCL frozen BERT variant, but only when using pre-initialized keys. 

When using incremental key initialization, performance was consistently subpar, indicating that DKVB requires access to a general-purpose corpus or the full input distribution to achieve competitive performance. 
While having access to the full data distribution may be unrealistic in practice, our experiments show that this is not needed in NLP. 
Rather, when initializing the keys using a general-purpose corpus, we obtain results that are close to the Oracle setup.

The largest performance drop between the DKVB variants and other CL methods was seen in the DIL setting. 
This suggests that DKVB's strength in preventing catastrophic forgetting through distinct key-value bindings becomes its weakness in DIL, as this compartmentalization restricts the model's ability to transfer knowledge across different domains.
Notably, NCL methods achieve similar results as the CL methods in this DIL setup, indicating that pre-trained language models without a bottleneck are already well-suited for domain incremental learning.

In the CIL and TIL tasks, only the frozen BERT NCL variant showcased performance comparable to that of the CL methods. 
The strong performance of frozen BERT in these experiments suggests that if task-ID is available during testing, a simple multi-head configuration with a frozen encoder is often sufficient.
Experiments in the single-head CIL setup have shown to be more challenging. 
As the models are tested on the full test set after each increment, ideally, they should exhibit a progressive increase in accuracy. 
But when no task-ID is provided and decoding is done with a single head, most models overfit and suffer catastrophic forgetting between increments, with DKVB being the only model to demonstrate improved CL capability in this scenario. 
This suggests that DKVB's unique architecture effectively maintains knowledge across tasks without needing task-specific heads.

\section{Conclusion}
\label{sec:conclusion}
The discrete key-value bottleneck offers an efficient approach to continual learning. 
It enables context-dependent updates in the model without explicit parameter isolation or dynamically expanding the architecture. 
Considering the special challenges of continual learning with text embeddings, we analyzed twelve architectural variants of the bottleneck. 
The best variants apply mean pooling after the bottleneck and utilize the hidden dimension of the encoded input representation to create the bottleneck heads.

We conducted a comprehensive evaluation across different continual learning settings in NLP, i.\,e., domain-incremental learning, class-incremental learning, and task-type incremental learning, and showed that with proper key initialization, the discrete key-value bottleneck offers consistent improvement in most settings and is comparable to dedicated continual learning methods from the literature.
Moreover, we showed that it can be used even in the most challenging single-head continual learning scenarios when no task-ID is provided.

\paragraph{Acknowledgments}
This research is co-funded by the CodeInspector project (No. 504226141) of the DFG, German Research Foundation.

\section{Limitations}
Our study focuses solely on encoder-only language models. 
While this raises questions about whether our results could generalize to other model architectures, our choice was motivated by their preference for supervised fine-tuning scenarios where the balance between performance and computational efficiency is crucial. Our experiments were also limited to fine-tuning for classification-based downstream tasks. 
Consequently, it remains to be investigated whether our results extend to other NLP tasks, such as semantic search, entity extraction, or machine translation.

\newpage

\bibliography{custom}
\bibliographystyle{acl_natbib}

\clearpage

\appendix

\section*{Supplementary Materials}

\section{Extended Related Work}
\label{sec:ext_related}

\paragraph{\textbf{Regularization-based methods}}
This family of methods involves incorporating explicit regularization terms to maintain a balance between the old and new tasks. This is usually done by adding penalty or regularization to the loss function to prevent large changes to parameters deemed important for old tasks~\citep{wang2024comprehensive}.
A popular method in this family is Elastic Weight Consideration (EWC), which calculates the importance of parameters with the Fisher information matrix, and applies smaller updates to weights deemed critical for earlier tasks~\citep{kirkpatrick2017overcoming}.

\paragraph{\textbf{Replay-based methods}}
These methods either retain a subset of training examples from previous tasks in memory such as A-GEM~\citep{chaudhry2018efficient}, or learn to generate pseudo samples from previous tasks, like in DGR~\citep{shin2017continual}. These samples are then incorporated into the training regimen of new tasks. While this can alleviate catasthropical forgetting, the size of a memory buffer is limited, which can potentially affect generalizability~\citep{wang2024comprehensive}.

\paragraph{\textbf{Optimization-based methods}} Explicitly manipulating the optimization process is another way to tackle the challenges of continual learning. Gradient-projection methods 
ensure that gradient updates happen exclusively in the orthogonal direction to the gradients of an old tasks, thereby preventing any impact on weights important for old tasks~\citep{zeng2019continual, guo2022adaptive}. Meta learning strategies and methods focusing on obtaining flat minima in the loss landscape can be also utilized in continual learning~\citep{javed2019meta, mirzadeh2020linear}.

\paragraph{\textbf{Architecture-based methods}} Methods in this family can be generally divided into \textit{parameter isolation} and, \textit{dynamic architecture} approaches, depending on whether the model architecture is fixed or not~\citep{wang2024comprehensive}. Models such as SupSup~\citep{wortsman2020supermasks} and HAT~\citep{serra2018overcoming} optimize a binary mask to selectively choose dedicated parameters for each task and fall under the parameter isolation category. Other methods dynamically expand the model with new parameters to increase capacity for learning new tasks~\citep{ke2021achieving, hung2019compacting}.

\paragraph{\textbf{Instruction-based methods}}
This family is unique to the field of NLP. These methods are based on task-specific instructions given to encoder-decoder or decoder only language models when a new task is encountered. While some methods in this family show promising knowledge transfer capabilities~\citep{scialom-etal-2022-fine, yin2022contintin, razdaibiedinaprogressive}, without explicit fine-tuning they are mostly limited by the knowledge acquired in the pre-training phase. 

\section{Extended Experimental Setup}
\label{sec:ext_setup}
\paragraph{Implementation}
For all model backbones in our experiments, we use the BERT-base model from Huggingface\footnote{https://huggingface.co/bert-base-uncased} and use cross-entropy loss as our objective function. We base our discrete-key-value bottleneck implementation on the \textit{vector-quantize-pytorch}\footnote{https://github.com/lucidrains/vector-quantize-pytorch} package. In the pre-experiments, we truncate each input sample to 256 tokens. For the main continual learning experiments, we rely on the \textit{PyContinual}\footnote{https://github.com/ZixuanKe/PyContinual} framework and reuse their implementations and hyperparameters on the baseline methods. To remain comparable to other studies using the \textit{PyContinual} framework, we kept the default preprocessing steps, used a maximum token length of 128, and applied the default convolutional decoder of the baseline models. For the single-head class incremental learning experiments we use a fixed randomized sequence order when creating the increments, and used a token length of 256. The source code for our experiments alongside the models can be found at \href{https://github.com/drndr/dkvb\_nlp}{github.com/drndr/dkvb\_nlp}.

\paragraph{Optimization}
As part of our pre-experiments, we also conducted a hyperparameter search and a sensitivity analysis on the bottleneck parameters. Outside of the selected hyperparameters and bottleneck parameters, all other configurations remained fixed during the search. Our experiments use the BERT-base architecture with a hidden size of 768. For the optimizer, we chose AdamW with a weight decay of 0.01. The dropout rate for the parametric decoder was set to 0.1. For the reference fully fine-tuned BERT numbers we reused the hyperparameters reported in \cite{galke2022bag}, for the frozen BERT variant we relied on the parametric DKVB variant hyperparameters with mean pooling.
During fine-tuning, we carefully optimized the models on both
datasets using grid-search-based manual tuning. A search space for the selected hyperparameters was defined, specifically we chose the batch size from $\{8, 16, 32\}$, the number of epochs from $\{5, 10\}$, the learning rate for the values layer from $\{1e\text{-}1, 1e\text{-}2, 1e\text{-}3\}$, and the decoder learning rate from $\{1e\text{-}3, 1e\text{-}4, 1e\text{-}5\}$. The best performing (based on the validation loss) configurations for each architecture variant can be seen in Table~\ref{tab:hyperparams}. The hyperparameters were reused for the continual learning main experiments.

In the single-head class incremental learning experiments we conducted an additional manual hyperparameter search for the DKVB variants and found a batch size of 16 with a global learning rate of $1e-2$ to be the best performing. For EWC we set the lambda parameter to 5,000. In the DER++ model we used a memory buffer size of 256 and set the sampling reate in each increment to 16.

\begin{table*}[h!]
    \scriptsize
    \centering
    \caption{Best hyperparameter configuration for each architecture variant based on validation loss \\}
    \begin{tabular}{lll|rrrr|rrrr}
         \toprule
         & & & \multicolumn{4}{c}{(A) Dataset: R8} & \multicolumn{4}{c}{(B) Dataset: 20ng} \\
         Model & Segmentation & Pooling & \#Epoch & Batch size & Values LR & Decoder LR &  \#Epoch & Batch size & Values LR & Decoder LR \\
         \midrule
         DKVB-P & hidden & Before CLS & 5 & 32 & $1e\text{-}2$ & $1e\text{-}4$ & 10 & 16 & $1e\text{-}1$ & $1e\text{-}3$ \\
         DKVB-P & hidden & Before Mean & 5 & 32 & $1e\text{-}1$ & $1e\text{-}4$ & 10 & 16 & $1e\text{-}2$ & $1e\text{-}4$ \\
         DKVB-P & hidden & After CLS & 10 & 16 & $1e\text{-}2$ & $1e\text{-}4$ & 5 & 16 & $1e\text{-}2$ & $1e\text{-}4$ \\
         DKVB-P & hidden & After Mean & 10 & 16 & $1e\text{-}2$ & $1e\text{-}3$ & 5 & 16 & $1e\text{-}2$ & $1e\text{-}3$ \\
         DKVB-P & token  & After CLS & 10 & 16 & $1e\text{-}2$ & $1e\text{-}3$ & 10 & 16 & $1e\text{-}2$ & $1e\text{-}3$ \\
         DKVB-P & token  & After Mean & 10 & 16 & $1e\text{-}2$ & $1e\text{-}3$ & 10 & 16 & $1e\text{-}2$ & $1e\text{-}3$ \\
         DKVB-NP & hidden & Before CLS & 5 & 32 & $1e\text{-}1$ & - & 5 & 32 & $1e\text{-}1$ & - \\
         DKVB-NP & hidden & Before Mean & 5 & 32 & $1e\text{-}1$ & - & 10 & 32 & $1e\text{-}1$ & - \\
         DKVB-NP & hidden & After CLS & 10 & 16 & $1e\text{-}1$ & - & 5 & 16 & $1e\text{-}1$ & - \\
         DKVB-NP & hidden & After Mean & 10 & 32 & $1e\text{-}1$ & - & 10 & 16 & $1e\text{-}1$ & - \\
         DKVB-NP & token  & After CLS & 10 & 32 & $1e\text{-}1$ & - & 10 & 32 & $1e\text{-}2$ & - \\
         DKVB-NP & token  & After Mean & 10 & 32 & $1e\text{-}1$ & - & 10 & 32 & $1e\text{-}2$ & -\\
         \bottomrule
    \end{tabular}
    \label{tab:hyperparams}
\end{table*}

\subsection{Bottleneck Parameters}
\label{ref:bottleneck_params}
In our experiments, we rely on the optimal bottleneck parameter analysis of \cite{trauble2022discrete}. Additionally, we also conduct a small sensitivity study for the discrete key dimension and number of key-value pairs on the R8 dataset. For this, we use the DKVB-NP model variant from the pre-experiments and keep everything fixed, changing only these two bottleneck parameters. For the base hyperparameters, we reuse the best-performing configurations.

\begin{figure}[h!]
    \centering
    \subfigure[]{\includegraphics[scale=0.305]{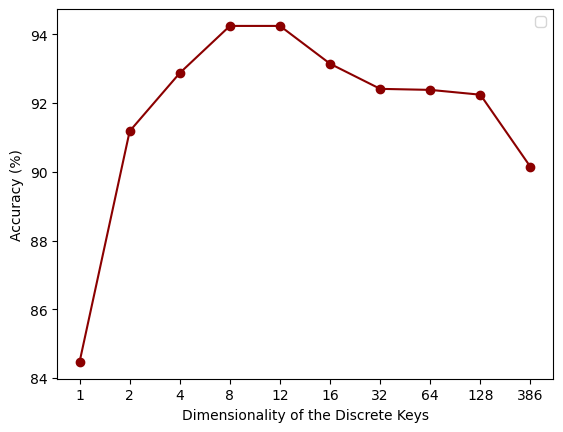}}
    \subfigure[]{\includegraphics[scale=0.305]{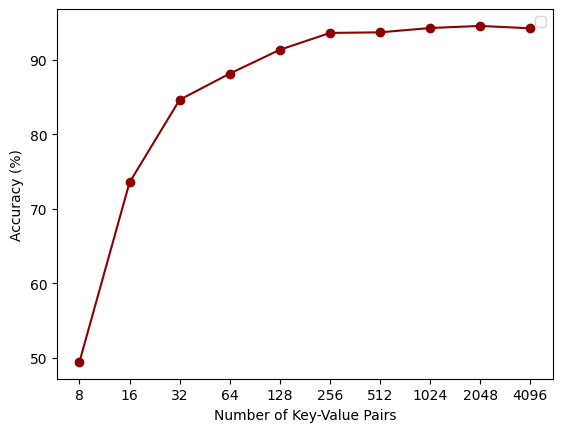}}
    \caption{Assessing the sensitivity of bottleneck parameters in regards of test accuracy: (a) Dimensionality of discrete key (b) Number of key-value pairs}
    \label{fig:sensitivity}
\end{figure}

\paragraph{Key dimension}
The number of dimensions of the discrete keys strongly influences the utility of the bottleneck. This can be explained as follows. Keys that have too few dimensions increase the chance of unintended key sharing between inputs from different distributions, while discrete keys with too high dimensionality can lead to insufficient coverage of the embedding space. Similarly to \cite{trauble2022discrete} we found the optimal key dimension to be between 8 to 12. The results of this analysis are depicted in Figure \ref{fig:sensitivity}.

\paragraph{Number of key-value pairs}
The number of key-value pairs determines the size of the discretized representational space. In accordance with the analysis of \cite{trauble2022discrete}, we found that increasing the number of key-value pairs leads to a performance increase. 
Eventually further increments no longer yield substantial improvements in performance. Note that increasing this parameter also leads to increased model size and increases the computational costs of key initialization as well. 
The results of this analysis are depicted in Figure \ref{fig:sensitivity}.

\section{Extended Results}
\label{sec:extended_exp}

\paragraph{Architectural Variants}
Results obtained using the RoBERTa (Table \ref{tab:roberta_res}) and DistilBERT (Table \ref{tab:distilbert_res}) models demonstrate comparable performance patterns to those observed with BERT. Interestingly, DistilBERT produced slightly better accuracy on most architecture variants compared to the other two models. 
However the highest performance was consistently seen with mean pooling after the bottleneck on all three models. DistilBERT's improved performance can likely be attributed to differences in its tokenization and pooling implementation compared to other models.

\begin{table*}[ht!]
    \scriptsize
    \centering
    \caption{Accuracy and standard deviation (in subscript) of the different DKVB architecture variants with RoBERTa on the R8 and 20ng datasets in a non-continual, standard learning setup, averaged over 5 runs.}
    \begin{tabular}{llcccc|cccc}
        \toprule
          & & \multicolumn{4}{c}{\textbf{Dataset: R8}} & \multicolumn{4}{c}{\textbf{Dataset: 20ng}} \\
         \midrule
          \multirow{2}{*}{\textbf{Decoder}} & \multirow{2}{*}{\textbf{Segmentation}} & \multicolumn{2}{c}{\textbf{Pooling Before}} & \multicolumn{2}{c}{\textbf{Pooling After}} & \multicolumn{2}{c}{\textbf{Pooling Before }} & \multicolumn{2}{c}{\textbf{Pooling After}}\\
        & & \textbf{CLS} & \textbf{Mean} & \textbf{CLS} & \textbf{Mean} & \textbf{CLS} & \textbf{Mean} & \textbf{CLS} & \textbf{Mean}\\
         \midrule
         \multirow{2}{*}{Parametric} & hidden & $49.48_{\textit{0.05}}$ & $91.36_{\textit{0.40}}$ & $91.73_{\textit{0.81}}$ & $\mathbf{94.25_{\textit{0.26}}}$ & $51.05_{\textit{0.43}}$ & $56.63_{\textit{0.39}}$ & $52.18_{\textit{1.11}}$ & $\mathbf{75.08_{\textit{0.21}}}$ \\
          & token  & - & - & $90.02_{\textit{1.05}}$ & $94.05_{\textit{0.31}}$ & - & - & $19.86_{\textit{1.03}}$ & $27.30_{\textit{0.92}}$ \\         
         \hline
         \multirow{2}{*}{Non Parametric} & hidden & $49.45_{\textit{0.02}}$ & $92.05_{\textit{0.29}}$ & $74.53_{\textit{1.74}}$ & $93.04_{\textit{0.20}}$ & $56.83_{\textit{0.35}}$ & $60.42_{\textit{0.35}}$ & $53.10_{\textit{1.37}} $ & $70.33_{\textit{0.78}}$ \\
         & token & - & - & $58.74_{\textit{0.92}}$ & $66.33_{\textit{0.74}}$ & - & - & $9.89_{\textit{0.66}}$ & $12.51_{\textit{1.67}}$ \\
         \midrule
         \midrule
         \multicolumn{2}{l}{RoBERTa (frozen) w/o DKVB} &
         \multicolumn{4}{c}{$94.29_{\textit{0.17}}$} & \multicolumn{4}{c}{$69.42_{\textit{0.30}}$} \\
         \multicolumn{2}{l}{RoBERTa w/o DKVB} & \multicolumn{4}{c}{$97.54_{\textit{0.51}}$} & \multicolumn{4}{c}{$83.36_{\textit{0.30}}$}\\
         \bottomrule
    \end{tabular}
    \label{tab:roberta_res}
\end{table*}

\begin{table*}[ht!]
    \scriptsize
    \centering
    \caption{Accuracy and standard deviation (in subscript) of the different DKVB architecture variants with DistilBERT on the R8 and 20ng datasets in a non-continual, standard learning setup, averaged over 5 runs.}
    \begin{tabular}{llcccc|cccc}
        \toprule
          & & \multicolumn{4}{c}{\textbf{Dataset: R8}} & \multicolumn{4}{c}{\textbf{Dataset: 20ng}} \\
         \midrule
          \multirow{2}{*}{\textbf{Decoder}} & \multirow{2}{*}{\textbf{Segmentation}} & \multicolumn{2}{c}{\textbf{Pooling Before}} & \multicolumn{2}{c}{\textbf{Pooling After}} & \multicolumn{2}{c}{\textbf{Pooling Before }} & \multicolumn{2}{c}{\textbf{Pooling After}}\\
        & & \textbf{CLS} & \textbf{Mean} & \textbf{CLS} & \textbf{Mean} & \textbf{CLS} & \textbf{Mean} & \textbf{CLS} & \textbf{Mean}\\
         \midrule
         \multirow{2}{*}{Parametric} & hidden & $90.22_{\textit{0.30}}$ & $92.17_{\textit{0.25}}$ & $90.46_{\textit{0.44}}$ & $\mathbf{96.09_{\textit{0.21}}}$ & $56.26_{\textit{0.53}}$ & $60.24_{\textit{0.20}}$ & $60.26_{\textit{0.31}}$ & $\mathbf{79.79_{\textit{0.49}}}$ \\
          & token  & - & - & $89.24_{\textit{0.93}}$ & $94.78_{\textit{0.86}}$ & - & - & $45.45_{\textit{1.13}}$ & $68.06_{\textit{0.90}}$ \\         
         \hline
         \multirow{2}{*}{Non Parametric} & hidden & $89.79_{\textit{0.24}}$ & $92.02_{\textit{0.38}}$ & $90.92_{\textit{0.22}}$ & $95.09_{\textit{0.27}}$ & $60.73_{\textit{0.48}}$ & $59.98_{\textit{0.30}}$ & $61.43_{\textit{0.50}} $ & $75.11_{\textit{0.44}}$ \\
         & token & - & - & $66.37_{\textit{0.49}}$ & $72.65_{\textit{0.24}}$ & - & - & $12.04_{\textit{0.51}}$ & $18.70_{\textit{1.03}}$ \\
         \midrule
         \midrule
         \multicolumn{2}{l}{DistilBERT (frozen) w/o DKVB} &
         \multicolumn{4}{c}{$94.62_{\textit{0.16}}$} & \multicolumn{4}{c}{$68.56_{\textit{0.38}}$} \\
         \multicolumn{2}{l}{DistilBERT w/o DKVB} & \multicolumn{4}{c}{$97.83_{\textit{0.24}}$} & \multicolumn{4}{c}{$83.84_{\textit{0.23}}$}\\
         \bottomrule
    \end{tabular}
    \label{tab:distilbert_res}
\end{table*}

\paragraph{Continual Learning Experiments} In the field of continual learning additional metrics are often used to measure the performance over incremental learning. Two often used metrics are Forward Transfer (FWT) and Backward Transfer (BWT) \cite{lopez2017gradient}. BWT refers to how learning a new task affects performance on a previously learned task. It can be positive, when learning a new task improves performance on the earlier task, or negative, when it worsens it. Severe negative backward transfer is often called catastrophic forgetting.
FWT describes how learning a new task influences performance on a future task. Since pre-trained language models already posses high transfer learning capabilities \cite{brown2020language}, its difficult to isolate the effect of learning specific task on future performance. Therefore we focus on BWT which is formally defined as:

\begin{equation}
\text{BWT} = \frac{1}{T - 1} \sum_{i=1}^{T-1} (R_{T,i} - R_{i,i})
\end{equation}

where $R \in \mathbb{R}^{T \times T}$ is the results matrix of an incremental learning scenario with $T$ tasks, where each entry $R_{i,j}$ being the test accuracy on task $j$ after training on task $i$ \cite{lopez2017gradient}.We report the BWT numbers on the three continual learning scenarios in Table \ref{tab:bwt}.

\begin{table*}[ht!]
    \scriptsize
    \centering
    \caption{Average Backward Transfer (BWT) scores on the three continual learning scenarios}
    \begin{tabular}{ll|rrr}
    \toprule
    \textbf{CL Method} &  \textbf{Model} & \textbf{DIL} & \textbf{CIL} & \textbf{TIL} \\
    \textbf{ } &  \textbf{  } & \textbf{(DSC)} & \textbf{(20NG)} & \textbf{(4GLUE)} \\
    \midrule
    NCL & BERT & $0.29$ & $-29.77$ & $-20.00$ \\
    NCL & BERT (frozen) & $-0.10$ & $-0.38$ & $-6.70$ \\
    NCL & Adapter-BERT  & $0.39$ & $-20.01$ & $-16.05$\\
    \hline
    DER++ & BERT (frozen) & $-0.58$ & $-27.11$ & $-7.05$ \\
    EWC & BERT (frozen) & $0.06$ & $\textbf{-0.27}$ &  $-10.84$ \\
    OWM & BERT (frozen) & $0.26$ & $-15.44$ & $-8.23$ \\
    CTR & Adapter-BERT & $\textbf{0.49}$ & $-0.50$ & $\textbf{-6.19}$ \\
    \midrule
    \midrule
    DKVB-NP Incremental& BERT (frozen) & $-3.06$ & $-27.73$ & $-21.30$ \\
    DKVB-NP Oracle & BERT (frozen) & $\textbf{-0.88}$  & $\textbf{-0.12}$ & $\textbf{-7.22}$  \\
    DKVB-NP Generic & BERT (frozen) & $-1.17$ & $-0.29$ & $-7.97$  \\
    DKVB-P Incremental& BERT (frozen) & $-4.88$ & $-29.05$ & $-20.99$  \\
    DKVB-P Oracle & BERT (frozen) & $-1.02$ & $-0.41$ & $-20.56$  \\
    DKVB-P Generic & BERT (frozen) & $-6.24$ & $-4.94$ & $-16.00$ \\
    \bottomrule
    \end{tabular}
    \label{tab:bwt}
\end{table*}

\begin{table*}[!h]
    \scriptsize
    \centering
    \caption{Mean accuracy scores of single-head class incremental learning experiments on R8, averaged over 5 runs with fixed sequence order\\}
    \begin{tabular}{ll|rrrrrrr}
    \toprule
       Increment & \vtop{\hbox{\strut \# Test}\hbox{\strut Samples}}  & BERT & BERT-frozen & \vtop{\hbox{\strut BERT-frozen}\hbox{\strut DER++}} & \vtop{\hbox{\strut BERT-frozen}\hbox{\strut EWC}}& \vtop{\hbox{\strut DKVB-NP}\hbox{\strut Incremental}} &
       \vtop{\hbox{\strut DKVB-NP}\hbox{\strut Oracle}} &
       \vtop{\hbox{\strut DKVB-NP}\hbox{\strut Wiki}} \\
       \midrule
       1. & 1596  & 31.79 & 31.79 & 31.79 & 31.79 & 31.79 & 31.79 & 31.79 \\
       2. & 253 & 5.52 & 5.52 & 8.60 & 31.79 & 5.57 & 31.80 & 31.79  \\
       3. & 2840 & 49.47 & 49.47 & 46.47 & 12.65 & 49.70 & 67.70 & 43.19 \\
       4. & 41 & 0.45 & 0.45 & 4.99 & 49.61 & 27.40 & 76.98 & 75.83 \\
       5. & 190 & 3.70 & 3.70 & 14.57 & 49.56 & 41.02 & 77.57 & 77.56 \\
       6. & 206 & 3.97 & 3.97 & 23.11 & 3.70 & 44.86 & 79.26 & 79.76 \\
       7. & 108 & 1.64 & 1.64 & 14.89 & 3,74 & 43.85 & 79.72 & 80.61 \\
       8. & 251 & 3.42 & 3.42 & 16.71 & 2.64 & 29.83 & 80.86 & 81.90 \\
    \end{tabular}
    \label{tab:class_incr_r8}
\end{table*}

\begin{table*}[h!]
    \scriptsize
    \centering
    \caption{Mean accuracy scores of single-head class incremental learning experiments on R52, averaged over 5 runs with fixed sequence order\\}
    \begin{tabular}{ll|rrrrrrrr}
    \toprule
       Increment & \vtop{\hbox{\strut \# Test}\hbox{\strut Samples}} & BERT & BERT-frozen & \vtop{\hbox{\strut BERT-frozen}\hbox{\strut DER++}} & \vtop{\hbox{\strut BERT-frozen}\hbox{\strut EWC}}& \vtop{\hbox{\strut DKVB-NP}\hbox{\strut Incremental}} & \vtop{\hbox{\strut DKVB-NP}\hbox{\strut Oracle}} &
       \vtop{\hbox{\strut DKVB-NP}\hbox{\strut Wiki}} \\
       \midrule
       1. & 45 & 0.70 & 0.50 & 0.56 & 0.50 & 0.73 & 0.54 & 0.60\\
       2. & 1600 & 26.20 & 27.10 & 26.07 & 0.50 & 27.10 & 27.10 & 0.92  \\
       3. & 52 & 0.46 & 0.23 & 14.99 & 0.85 & 0.54 & 27.29 & 24.71 \\
       4. & 29 & 0.35 & 0.35 & 9.60 & 3.69 & 0.46 & 29.51 & 27.30 \\
       5. & 321 & 2.92 & 2.95 & 4.08 & 2.57 & 3.58 & 25.60 & 27.41 \\
       6. & 37 & 0.35 & 0.07 & 18.76 & 0.35 & 0.42 & 27.22 & 27.63\\
       7. & 17 & 0.35 & 0.42 & 7.73 & 0.35 & 0.35 & 28.85 &  27.55\\
       8. & 44 & 0.54 & 0.46 & 4.53 & 0.35 & 0.70 & 27.57 &  27.67\\
       9. & 28 & 0.50 & 0.35 & 13.13 & 0.42 & 0.46 & 27.10 &  28.10\\
       10. & 110 & 1.40 & 1.40 & 12.73 & 0.35 & 1.40 & 25.58 & 28.56 \\
       11. & 3046 & 42.17 & 42.17 & 44.18 & 0.35 & 44.82 & 42.17 & 28.87\\
       12. & 16 & 0.97 & 0.35 & 19.78 & 0.70 & 0.42 & 43.71 &  29.17\\
       13. & 10 & 0.15 & 0.23 & 25.31 & 0.97 & 0.35 & 44.85 &  29.48\\
       14. & 193 & 3.15 & 3.15 & 21.58 & 0.42 & 3.15 & 52.64 & 30.30 \\
       15. & 213 & 3.38 & 3.38 & 23.62 & 3.38 & 2.95 & 42.52 &  31.80\\
       16. & 154 & 1.09 & 1.09 & 6.43 & 3.38 & 1.55 & 42.83 &  35.83\\
       17. & 145 & 1.40 & 1.40 & 26.11 & 1.47 & 1.83 & 45.40 & 36.12 \\
       18. & 32 & 0.50 & 0.50 & 23.19 & 0.50 & 0.50 & 45.52 &  17.95\\
       19. & 203 & 3.15 & 3.15 & 31.43 & 1.83 & 3.30 & 45.71 & 11.10 \\
       20. & 227 & 3.15 & 3.15 & 25.42 & 3.15 & 3.62 & 38.94 &  10.34\\
       21. & 2948 & 42.17 & 42.17 & 45.67 & 3.15 & 42.25 & 44.74 & 40.07 \\
       22. & 255 & 4.71 & 4.71 & 47.63 & 42.25 & 4.71 & 48.84 & 41.96 \\
       23. & 59 & 0.58 & 0.38 & 23.27 & 42.17 & 0.77 & 47.15 & 42.23 \\
       24. & 48 & 0.58 & 0.58 & 22.85 & 0.58 & 0.62 & 37.96 &  42.48\\
       25. & 59 & 0.58 & 0.58 & 42.04 &  0.58 & 0.70 & 38.55 & 43.11 \\
       26. & 243 & 3.38 & 3.38 & 35.75 & 0.58 & 3.62 & 47.78 &  45.04 \\
   \end{tabular}
    \label{tab:class_incr_r52}
\end{table*}
\end{document}